\documentclass[10pt,twocolumn,letterpaper]{article}

\usepackage{iccv}              

\usepackage[misc]{ifsym}  
\usepackage{colortbl}  
\usepackage{multirow}

\definecolor{cvprblue}{rgb}{0.21,0.49,0.74}
\usepackage[pagebackref,breaklinks,colorlinks,allcolors=cvprblue]{hyperref}

\def\paperID{7} 
\def\confName{ICCV}
\def\confYear{2025}

\title{MGT: Extending Virtual Try-Off to Multi-Garment Scenarios}

\author{Riza Velioglu\textsuperscript{\Letter}, Petra Bevandic, Robin Chan, Barbara Hammer\\
Machine Learning Group, CITEC, Bielefeld University, Germany\\
{\tt\small \{rvelioglu, pbevandic, rchan, bhammer\}@techfak.de}
}

\begin{document}
\twocolumn[{%
\renewcommand\twocolumn[1][]{#1}%
\maketitle
\begin{center}
    \centering
    \captionsetup{type=figure}
        \includegraphics[width=.999\linewidth]{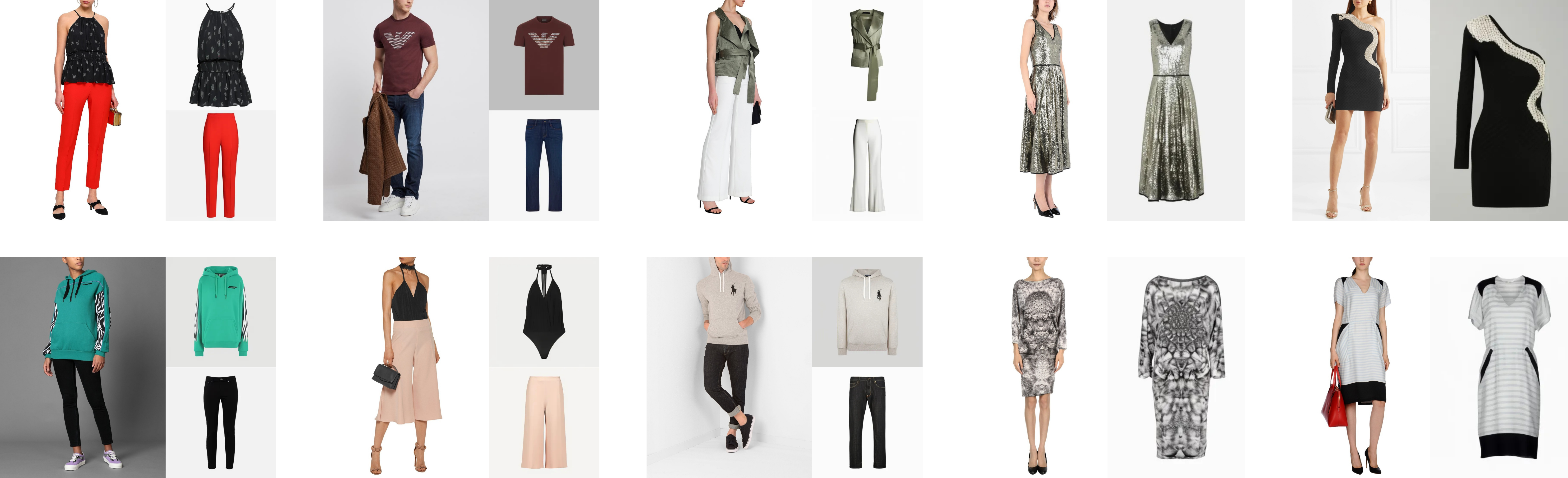}
    \captionof{figure}{\textbf{Virtual try-off results generated by Multi-Garment TryOffDiff~(MGT)}. 
    The first three columns demonstrate MGT's ability to generate multi-garment images (\eg, upper- and lower-body garments) from a single reference image. 
    The last two columns illustrate dress reconstruction results.
    MGT generates an image in under 3 seconds on a consumer GPU.
    Please zoom in for a clearer view of details.
    }
    \label{fig:cover}
\end{center}%
}]
\begin{abstract}

\vspace{-.6em}
Computer vision is transforming fashion industry through
Virtual Try-On (VTON) and Virtual Try-Off (VTOFF).
VTON generates images of a person in a specified garment using a target photo
and a standardized garment image, while
a more challenging variant, Person-to-Person Virtual Try-On (p2p-VTON), 
uses a photo of another person wearing the garment.
VTOFF, in contrast, extracts standardized garment images from photos of clothed individuals.
We introduce Multi-Garment TryOffDiff (MGT), a diffusion-based VTOFF model 
capable of handling diverse garment types, including upper-body, lower-body, and dresses.
MGT builds on a latent diffusion architecture with SigLIP-based image conditioning to capture garment characteristics such as shape, texture, and pattern.
To address garment diversity, MGT incorporates class-specific embeddings, 
achieving state-of-the-art VTOFF results on VITON-HD 
and competitive performance on DressCode.
When paired with VTON models, it further enhances p2p-VTON 
by reducing unwanted attribute transfer, such as skin tone, 
ensuring preservation of person-specific characteristics.
Demo, code, and models are available at: \url{https://rizavelioglu.github.io/tryoffdiff/}
\end{abstract}
    
\section{Introduction}\label{sec:introduction}

In the rapidly evolving landscape of e-commerce, particularly within the fashion industry, the ability to provide customers with immersive and personalized shopping experiences is crucial for brands to differentiate themselves and drive engagement. Virtual Try-On (VTON)~\citep{jetchev2017conditional} technology has emerged as a powerful tool in this regard, allowing customers to visualize how garments would look on them without physically trying them on. 
Traditional VTON systems rely on standardized product images from e-commerce catalogs~\citep{choi2021viton, morelli2022dress}, which may not fully capture the nuances of how a garment appears when worn by a real person, and are often unavailable for vintage, user‑generated, or unbranded items. These constraints can reduce the authenticity of the virtual try-on experience, potentially affecting customer satisfaction and conversion rates.

Virtual Try‑Off (VTOFF)~\cite{velioglu2024tryoffdiff} overcomes these limitations by reconstructing standardized garment images directly from photos of people wearing them, addressing the scarcity of standardized product shots.
Unlike VTON, which synthesizes a person wearing a specified garment, VTOFF extracts the garment itself in a clean, catalog-style format, free of stylistic variations. 
This capability is highly valuable in advertising and marketing, where visuals assets strongly influence purchasing behavior~\cite{xia2020creating,VANDERHEIDE2013570}. 
Traditional production of catalog imagery demands specialized equipment and extensive post‑processing, making it both time‑consuming and costly. By contrast, VTOFF enables rapid generation of consistent, high‑quality garment visuals without a dedicated studio setup, streamlining product promotion and boosting customer engagement. 
Furthermore, VTOFF supports visual retrieval from user-generated content, enabling automated trend analysis by isolating garments from influencer or street-style images, thus enriching brand intelligence and market research.

Moreover, VTOFF opens new possibilities for person-to-person Virtual Try-On (p2p-VTON)~\citep{xie2021towards}, a more challenging variant of VTON that uses a photo of another person wearing the desired garment as the conditioning input. By integrating VTOFF with traditional VTON pipelines, p2p-VTON becomes feasible without requiring standardized catalog images. This functionality is especially appealing in social commerce, where peer influence drives purchasing decisions. By leveraging VTOFF, brands can create interactive, socially engaging shopping experiences, boosting conversion rates and fostering stronger connections with their audiences.

In this paper, we present Multi-Garment TryOffDiff (MGT), a diffusion-based model designed to address the unique challenges of VTOFF task.
Built upon Stable Diffusion architecture, MGT replaces text conditioning with image conditioning, utilizing SigLIP features~\cite{zhai2023sigmoid} processed through an adapter module to capture garment-specific properties such as texture, shape and patterns. 
This approach yields competitive results on the VITON-HD dataset for upper-body garment reconstruction, despite not being trained on it, highlighting robust cross-domains generalization.
To handle multi-garment scenarios, we extend TryOffDiff~\cite{velioglu2024tryoffdiff} by incorporating class-specific embeddings into the timestep embeddings~\cite{nichol2021improved}.
Evaluated on the full DressCode dataset, MGT performs comparably to garment-specific models, establishing it as the first model to support multi-garment VTOFF.
Additionally, integrating MGT with state-of-the-art VTON models enhances p2p-VTON by minimizing unwanted attribute transfer, such as skin color, from the source person.
Our key contributions are:
\begin{itemize}
    \item \textbf{Multi-Garment TryOffDiff}: A diffusion-based virtual try-off model that introduces class-specific embeddings to enable simultaneous reconstruction of multiple garment categories, delivering robust cross-domain generalization and marking \emph{the first approach} to support multi-garment try-off scenarios.
    \item Integration of VTOFF outputs with existing try-on pipelines, enabling p2p-VTON with improved preservation of person-specific attributes and reduced unwanted attribute transfer, resulting in higher-fidelity outputs.
\end{itemize}

\begin{figure}[t]
    \centering
    \includegraphics[width=0.99\linewidth]{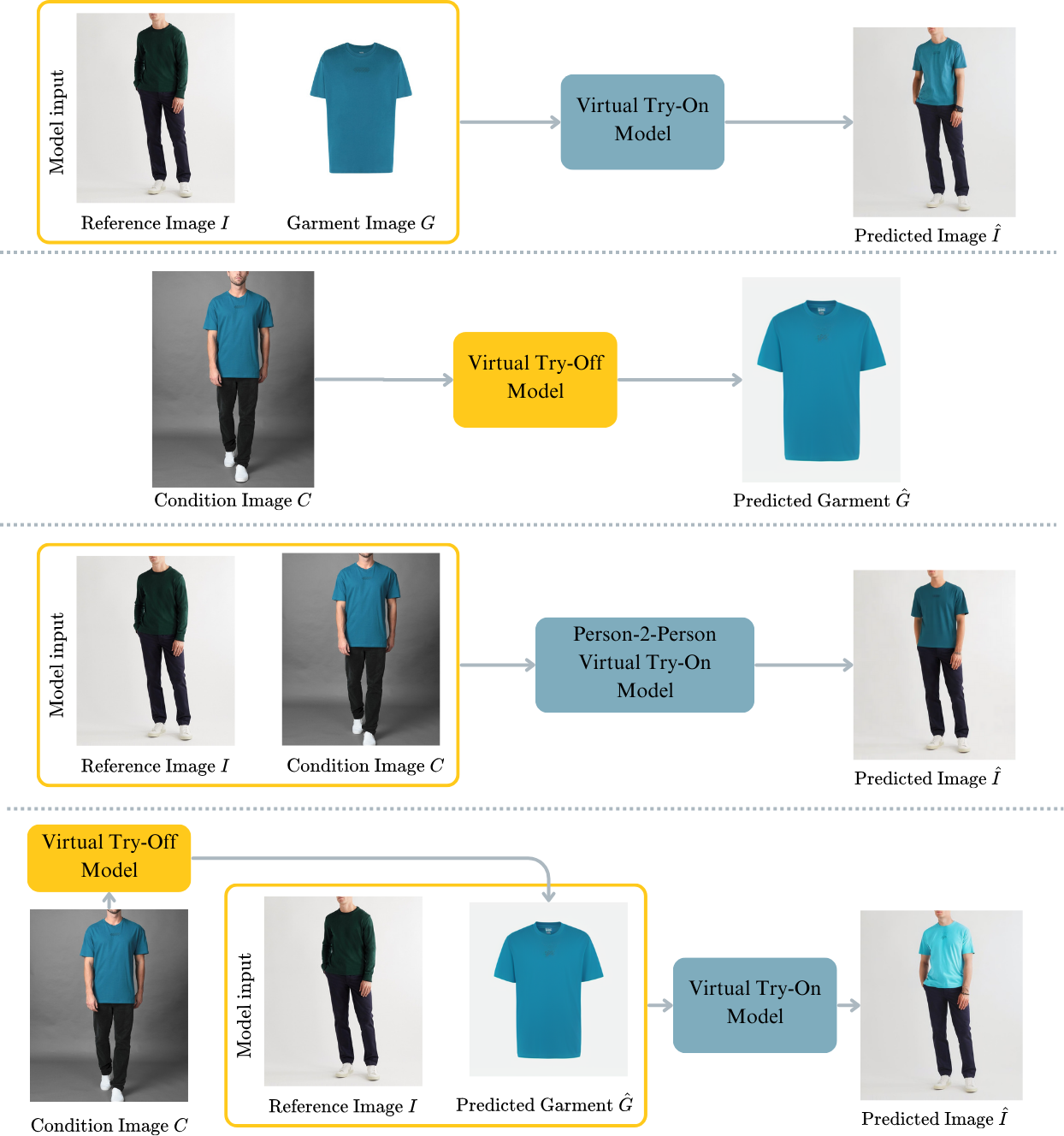}
    \caption{\textbf{Overview of various fashion image generation pipelines.}
    First row illustrates Virtual Try-On, which synthesizes an image of a person wearing a target garment given input images of person and garment.
    Second row depicts Virtual Try-Off, which generates an e-commerce-style image of a garment from a photo of a person wearing it.
    Third row shows Person-to-Person Virtual Try-On, where a garment is transferred from one person to another. 
    Finally, fourth row demonstrates that p2p-VTON can be achieved by integrating VTON and VTOFF models.}
    \label{fig:vtoff}
\end{figure}
\section{Related Work}
\label{sec:related_work}

This section reviews prior work on Virtual Try-Off, Image-based Virtual Try-On, and Conditional Diffusion Models. Virtual Try-Off generates standardized garment images from clothed individuals. Image-based Virtual Try-On synthesizes images of a person wearing a given garment. Conditional Diffusion Models serve as the generative backbone of recent approaches, supporting controlled synthesis through tailored conditioning mechanisms. These areas form the basis for our work on multi-garment reconstruction.

\paragraph{Virtual Try-Off.}
Virtual Try-Off~(VTOFF)~\cite{velioglu2024tryoffdiff} generates standardized garment images from photos of dressed people, a task with growing interest due to its applications in e-commerce.
Early works, such as TileGAN~\cite{zeng2020tilegan}, used a two-stage pipeline: a U-Net-like encoder-decoder for coarse synthesis followed by a pix2pix-based refinement.
With the advent of text-to-image diffusion models such as Stable Diffusion~\cite{rombach2022high}, recent works explored text-guided garment generation.
ARMANI~\cite{zhang2022armani}, DiffCloth~\cite{zhang2023diffcloth}, GarmentAligner~\cite{zhang2024garmentaligner}, and MGD~\cite{baldrati2023multimodal} finetuned latent diffusion models using garment-caption datasets.
However, text-based methods require detailed captions and often fail to follow them precisely, limiting quality.

To address this, image-based conditioning was introduced.
TryOffDiff~\cite{velioglu2024tryoffdiff} formalized the VTOFF task and replaced text embeddings in Stable Diffusion v1.4 with image embeddings.
IGR~\cite{shen2024igr} proposed a dual-tower variant of Stable Diffusion v1.5.
TryOffAnyone~\cite{xarchakos2024tryoffanyone} finetuned only self-attention layers, following CatVTON~\cite{chong2024catvton}, enabling a lightweight model with competitive results. 
Despite progress, VTOFF remains underexplored. 
Many models are not publicly available, and most focus solely on upper-body garments, limiting generalizability.

\paragraph{Image-based Virtual Try-On.}
Image-based virtual try-on (VTON) generates images of a person wearing a specific garment, preserving identity, pose, and body shape while accurately rendering garment details. 
CAGAN~\cite{jetchev2017conditional} introduced this with a cycle-GAN framework. 
VITON~\cite{han2018viton} formalized it as a two-step pipeline: warping the garment using non-parametric geometric transformations~\cite{belongie2002shape}, then blending it with the person image. 
CP-VTON~\cite{wang2018toward} introduced a learnable thin-plate spline~(TPS) transformation via a geometric matcher, later enhanced by dense flow~\cite{han2019clothflow} and appearance flow~\cite{ge2021parser} to better align textures and folds. However, warping-based methods struggle with occluded images as they lack generative capabilities.

Recent efforts shifted to GANs and diffusion models.
FW-GAN~\cite{dong2019fw} targeted try-on videos.
PASTA-GAN~\cite{xie2021towards} applied StyleGAN2 to person-to-person try-on.
GANs, however, suffer from instability and mode collapse, which diffusion models avoid.
IDM-VTON~\cite{choi2024improving} used two modules to encode garment semantics, extracting high- and low-level features with cross- and self-attention layers.
OOTDiffusion~\cite{xu2024ootdiffusion} leveraged pretrained latent diffusion models, integrating garment features into a denoising UNet via outfitting fusion.
In a lighter approach, CatVTON~\cite{chong2024catvton} avoided heavy feature extraction, proposing a compact model from a pretrained latent diffusion model with promising results using fewer parameters.

Adapting VTON models for VTOFF is ineffective,
as VTON often relies on additional inputs such as text prompts, keypoints, or segmentation masks, requiring careful adjustment~\cite{velioglu2024tryoffdiff}.
More importantly, while both VTON and VTOFF tasks involve garment manipulation, they differ fundamentally. VTON has access to complete garment information and focuses on adapting it to a given pose. In contrast, VTOFF must recover standardized garments from partially visible, occluded, or deformed inputs, requiring the model to infer missing details. 

\paragraph{Conditional Diffusion Models.}
Latent Diffusion Models~\cite{rombach2022high}~(LDMs)
generate high-quality images using cross-attention for conditioning on various modalities, including 
text~\cite{betker2023improving,baldridge2024imagen,esser2024scaling} and images~\cite{saharia2022palette,parmar2023zero,saharia2022image}. 
Text-guided methods such as ControlNet~\cite{zhang2023adding} and T2I-Adapter~\cite{mou2024t2i} improve spatial control with auxiliary networks. IP-Adapter~\cite{ye2023ipadapter} 
further improves flexibility by decoupling cross-attention for text and image features,
allowing image-guided generation.
Prompt-Free Diffusion~\cite{xu2024prompt} eliminates text prompts entirely, relying on reference image and optional structural inputs for guidance.

While effective for general image synthesis, existing conditional diffusion models are not well suited for garment reconstruction. 
Text-guided methods require highly detailed and consistent prompts to specify product attributes, which is labor-intensive and difficult to scale.
Image-guided approaches often lack the precision required for fashion photography~\cite{cheong2024visconet}. As shown in TryOffDiff~\cite{velioglu2024tryoffdiff}, applying these models directly to VTOFF results in low-fidelity outputs.

\section{Methodology}
\label{sec:methodology}

\begin{figure*}[t]
    \centering
    \includegraphics[width=0.99\linewidth]{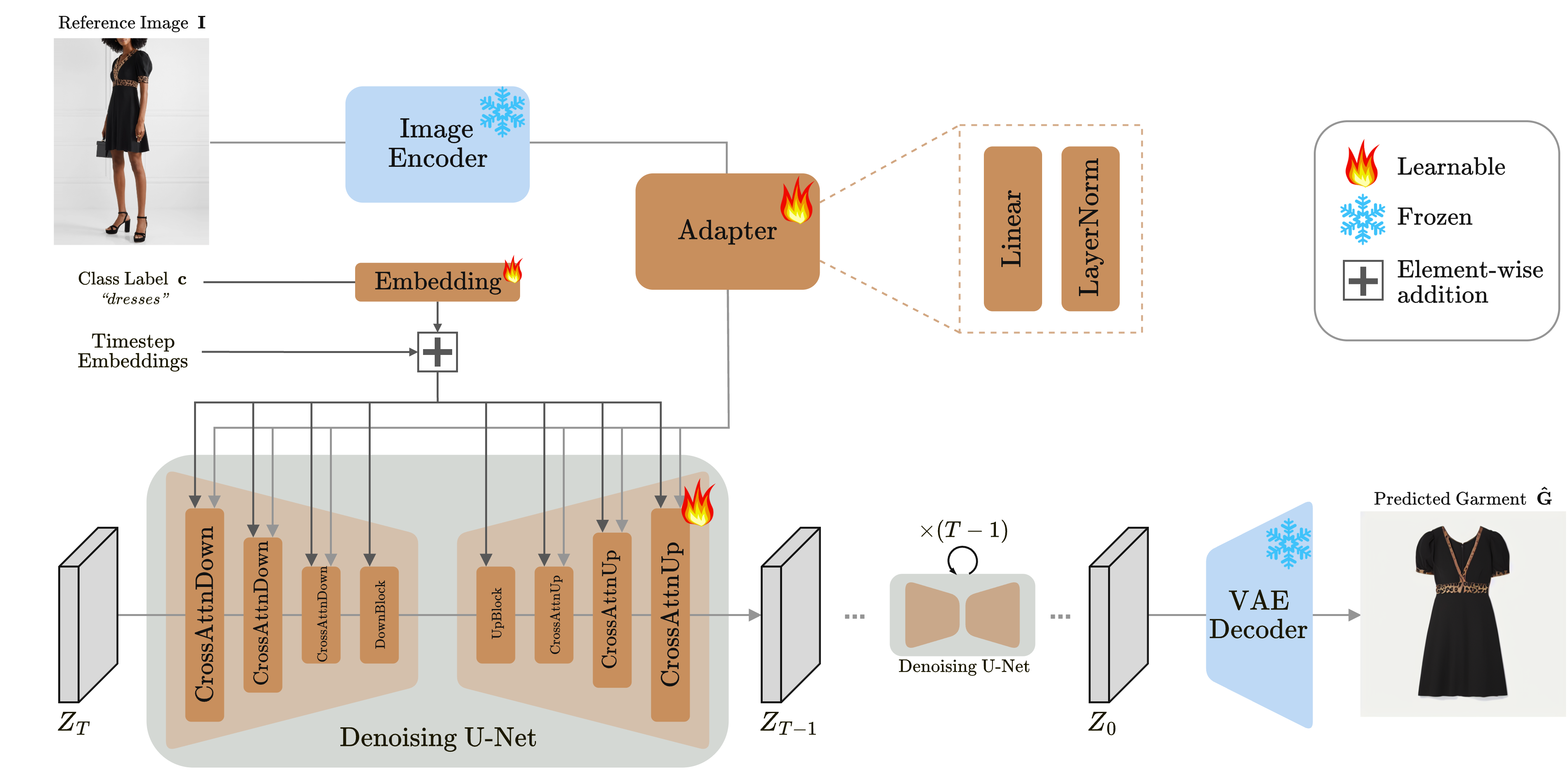}
    \caption{
        \textbf{Overview of MGT.}
         Given a reference image and a class label (\eg `dresses'), the SigLIP~\cite{zhai2023sigmoid} image encoder extracts features, which are subsequently processed by a learnable adapter and embedded into the cross-attention layers of the Denoising U-Net. 
         A learnable class label embedding conditions the generation to support multi-garment reconstruction. The model, with trainable Adapter, Embedding, and U-Net components, produces a standardized garment image, which is decoded by a frozen VAE.
        }
    \label{fig:TryOffDiff}
\end{figure*}

This section formally defines the virtual try-off task and outlines Multi-Garment TryOffDiff model.

\subsection{Virtual Try-Off}\label{sec:vtoff}
Consider an RGB image $\mathbf{I} \in \{0, \ldots, 255\}^{H \times W \times 3}$     
with height and width $H, W \in \mathbb{N}$, representing a person wearing garments. 
The VTOFF task aims to produce a standardized product
image $\mathbf{G} \in \{0, \ldots, 255\}^{H \times W \times 3}$, 
that meets commercial catalog specifications.
Formally, the goal is to train a generative model 
that learns the conditional distribution $P(G|C)$,
where $G$ and $C$ represent the variables 
corresponding to garment images
and reference images (serving as condition), respectively.
Suppose the model approximates this target distribution with $Q(G|C)$.
Then, given a specific reference image $\mathbf{I}$ as conditioning input, 
the objective is for a sample $\hat{\mathbf{G}} \sim Q(G | C = \mathbf{I})$
to resemble a true sample of a garment image $\mathbf{G} \sim P(G | C = \mathbf{I})$ 
as closely as possible.

\subsection{Multi-Garment TryOffDiff (MGT)} \label{sec:TryOffDiff}

The product image $\mathbf{G}$ may contain various types of garments, such as upper-body wear, lower-body wear, or full-body dresses. Reconstructing a specific garment type from full-body photos makes the virtual try-off task more challenging and necessitates additional guidance. To address this, we extend the existing VTOFF model TryOffDiff \cite{velioglu2024tryoffdiff} with class-specific embeddings that help disentangle and reconstruct individual garment categories.

\paragraph{Image Conditioning.}
A central challenge in image-guided generation
is effectively incorporating visual features
into the generative model's conditioning mechanism.
CLIP's ViT~\cite{radford2021learning} is a common choice for image encoding due to its general-purpose representations, but recent improvements by SigLIP~\cite{zhai2023sigmoid} make it better suited for detailed visual tasks. We adopt SigLIP as the image encoder and extract the full sequence of tokens from its final layer to retain spatial information. 
These token embeddings are projected and normalized (LN)~\cite{ba2016layer} via a lightweight adapter module:
\begin{equation}
\mathbf{C}(\mathbf{I}) = (\mathrm{LN} \circ \mathrm{Linear} \circ \mathrm{SigLIP})(\mathbf{I}) \in \mathbb{R}^{n \times m}
\end{equation}
where $\mathbf{C}(\mathbf{I})$ denotes the adapted feature sequence, and $n,m \in \mathbb{N}$ denote the token and feature dimensions, respectively. This design follows the IP-Adapter~\cite{ye2023ipadapter}.
The resulting features are injected into the cross-attention layers of the denoising U-Net.
Specifically, the key $\mathbf{K}$ and value $\mathbf{V}$ of the attention mechanism 
at each layer are derived from the image features through linear transformations:
\begin{equation}\label{eq:cross-attention-conditioning}
    \mathbf{K} = \mathbf{C}(\mathbf{I}) \cdot \mathbf{W}_k \in \mathbb{R}^{n \times d_k}, \mathbf{V} = \mathbf{C}(\mathbf{I}) \cdot \mathbf{W}_V \in \mathbb{R}^{n \times d_v}
\end{equation}
where $\mathbf{W}_k \in \mathbb{R}^{m \times d_k}$ and $\mathbf{W}_v \in \mathbb{R}^{m \times d_v}$.
This conditioning has already proven effective for VTOFF results, as demonstrated by TryOffDiff \cite{velioglu2024tryoffdiff}.

\paragraph{Garment Type Conditioning.}
Multi-Garment TryOffDiff (MGT) extends TryOffDiff~\cite{velioglu2024tryoffdiff} by introducing a simple class conditioning mechanism to support multiple garment types within a single model. 
Reference images often contain several garments (\eg, shirts and pants), requiring the model to distinguish which garment to reconstruct. To resolve this ambiguity and enable multi-garment support, we condition the model on garment category labels—``lower body'', ``upper body'', or ``dress''—consistent with the class annotations in the DressCode dataset. Each category is associated with a learnable embedding vector.

Let $\mathcal{C} = \{1, 2, 3\}$ denote the index set of garment classes and $\mathbf{E_c} \in \mathbb{R}^{|\mathcal{C}| \times d}$ be the learnable embedding matrix, where $d = 1280$ matches the diffusion model’s timestep embedding dimension. Given a class index $c \in \mathcal{C}$, the corresponding garment type embedding is retrieved as:
\begin{equation}
\mathbf{e}_c = \mathbf{E_c}[c] \in \mathbb{R}^d
\end{equation}
Let $\mathcal{T} = \{1, \cdots, T\}$, where $T=1000$, denote the index set of timesteps and $\mathbf{E_t} \in \mathbb{R}^{T \times d}$ be the learnable timestep embedding matrix. Given a timestep index $t \in \mathcal{T}$, the corresponding timestep embedding is retrieved as:
\begin{equation}
\mathbf{e}_t = \mathbf{E_t}[t] \in \mathbb{R}^d
\end{equation}
Then, the conditioned timestep embedding is defined as the element-wise addition of the timestep embedding and the garment type embedding:
\begin{equation}
    \mathbf{e_t}_{\text{cond}} = \mathbf{e}_t + \mathbf{e}_c
\end{equation}
This conditioned timestep embedding is injected into each residual block of the denoising U-Net~(following \cite{nichol2021improved}), where it modulates feature activations to guide generation towards the target garment class. 
This mechanism allows the model to disambiguate between garments types and produce semantically accurate outputs.

\section{Experiments}
\label{sec:experiments}

Since Virtual Try-Off is a newly introduced task, few open‐source methods exist. We evaluate our Multi-Garment TryOffDiff (MGT) against two publicly-available baselines: TryOffDiff~\cite{velioglu2024tryoffdiff} and TryOffAnyone~\cite{xarchakos2024tryoffanyone}, to establish benchmarks. 
We adopt DISTS~\cite{ding2020image} as our primary metric and report additional standard measures for a comprehensive evaluation. 
We also present qualitative examples to illustrate MGT’s performance on diverse inputs.

\subsection{Experimental Setup}
\label{sec:experiments_setup}

\paragraph{Datasets.}
Our experiments are conducted on two publicly available datasets:
VITON-HD~\cite{choi2021viton} and DressCode~\cite{morelli2022dress}.
Originally designed for VTON, these datasets are also well-suited for VTOFF as they provide image pairs $(\mathbf{I},\mathbf{G})$,
where $\mathbf{I}$ is a clothed person and $\mathbf{G}$ is the corresponding garment.

VITON-HD includes $13,679$ high-resolution~($1024 \times 768$) 
image pairs of frontal half-body models with upper-body garments. 
We preprocessed the dataset by removing duplicates
and test set leaks from the training set, resulting in $11,552$ unique pairs for training and $1,990$ for testing.

DressCode contains $53,792$ high-resolution~($1024 \times 768$) person-garment pairs, with $48,392$ pairs for training--comprising dresses~($27,678$), upper-body~($13,563$), and lower-body~($7,151$)--and $5,400$ pairs for testing, evenly split with $1,800$ pairs per category.

For person-to-person (p2p) try-on, where ground-truth are unavailable, we randomly pair garments across individuals and provide a text file of input image filenames for reproducibility. For DressCode, a label (upper-body, lower-body, or dresses) indicates the garment to transfer; this is not needed for VITON-HD, which includes only upper-body garments.

\paragraph{Evaluation Metrics.}
To evaluate reconstruction quality, we use established full-reference metrics such as 
\emph{Structural Similarity Index Measure} (SSIM)~\cite{wang2004image} and its \emph{multi-scale} (MS-SSIM) as well as \emph{complex-wavelet}~(CW-SSIM) variants.
To measure perceptual quality, we use common no-reference metrics like \emph{Learned Perceptual Image Patch Similarity} (LPIPS) \cite{zhang2018perceptual}, \emph{Fréchet Inception Distance} (FID)~\cite{heusel2017gans} and \emph{Kernel Inception Distance} (KID)~\cite{binkowski2018demystifying} metric.
To measure the perceptual similarity between images capturing both structural and textural information, we report the \emph{Deep Image Structure and Texture Similarity} (DISTS) \cite{ding2020image} metric.
We prioritize the DISTS metric in evaluations because it balances structural and textural information, offering a comprehensive evaluation of image quality.

\paragraph{Implementation Details.}
Denoising U-Net weights are initialized from Stable Diffusion v1.4~\cite{rombach2022high} and subsequently finetuned. The Adapter module and class Embedding layer are trained from scratch. The SigLIP encoder, VAE encoder, and VAE decoder are kept frozen throughout the training, which is performed end-to-end.
Input reference images are padded along the width to achieve a square aspect ratio and resized to $512 \times 512$ to match the pretrained SigLIP and VAE encoder's format.
Garment images undergo the same preprocessing during training. 
We use SigLIP-B/16-512 as image feature extractor, yielding 1,024 token embeddings of dimension 768, which the adapter--comprising a linear layer and normalization--reduces to $n=77$ conditioning embeddings of dimension $m=768$.
For garment conditioning, we embed class labels (\eg `upper-body', `lower-body', `dress') using a trainable embedding layer. This layer has 3 classes and projects the labels into a 1,280-dimensional space, matching the dimension of the timestep embeddings. These class embeddings are then combined with the timestep embeddings via element-wise addition before being fed into the U-Net’s residual blocks.

Training runs for 200 epochs (approximately 150k iterations) on a single node with four NVIDIA A40 GPUs, taking about 5 days with a batch size of 16. 
We use the AdamW optimizer~\cite{loshchilov2018decoupled}, 
with an initial learning rate of 5e-5,
which increases linearly from zero over the first 15,000 warmup steps (\%10 of total iterations), then decays linearly to zero. 
A weight decay of 0.01 applies to all parameters except biases and normalization weights. Following~\cite{karras2022elucidating}, 
we use the Euler noise scheduler~(Algorithm 2) with 1,000 steps.
Optimization relies on the standard Mean Squared Error (MSE) loss, 
which measures the distance between the added 
and the predicted noise at each step \cite{ho2020denoising}.

During inference, TryOffDiff uses Euler scheduler with 20 timesteps and a guidance scale of 1.5.
On a single NVIDIA A40 GPU, inference on DressCode test set takes 2.8 seconds per image and requires 10.1GB of memory.

\subsection{Quantitative Results}\label{seq:quantitative-results}

We compare MGT to TryOffDiff~\cite{velioglu2024tryoffdiff} and TryOffAnyone~\cite{xarchakos2024tryoffanyone} on the VITON-HD and DressCode datasets, with results reported in \Cref{tab:vtoff_dresscode} (DressCode), \Cref{tab:ablations_vitonhd} (VITON-HD), and \Cref{tab:p2p} (p2p-VTON).
TryOffDiff and TryOffAnyone are trained on VITON-HD (upper-body only), while MGT is trained on DressCode, supporting multiple garment categories (upper-body, lower-body, dresses).

\begin{table}
\setlength{\tabcolsep}{2.5pt}
\centering
\resizebox{\linewidth}{!}{
\begin{tabular}{lcccccccc}
    \toprule
    \textbf{Method} & SSIM$\uparrow$ & MS-$\uparrow$ & CW-$\uparrow$ & LPIPS$\downarrow$ & FID$\downarrow$ & FD$^\text{CLIP}$$\downarrow$ & KID$\downarrow$ & DISTS$\downarrow$ \\
    \midrule
    \rowcolor{gray!20}\multicolumn{9}{c}{Upper-body} \\
    TryOffDiff-single     & 80.8 & 73.8 & 47.8 & 31.6 & 17.1 & 5.2 & 4.7 & 21.6 \\
    MGT (ours)            & 80.2 & 73.2 & 48.1 & 32.3 & 17.8 & 5.3 & 5.5 & 22.2 \\
    \rowcolor{gray!20}\multicolumn{9}{c}{Lower-body} \\
    TryOffDiff-single     & 81.1 & 76.8 & 55.5 & 30.0 & 22.6 & 9.4 & 5.9 & 21.1 \\
    MGT (ours)            & 80.4 & 75.6 & 54.1 & 30.8 & 22.4 & 9.0 & 6.8 & 21.7 \\
    \rowcolor{gray!20}\multicolumn{9}{c}{Dresses} \\
    TryOffDiff-single     & 81.6 & 76.1 & 55.6 & 26.1 & 18.4 & 8.2 & 4.4 & 20.8 \\
    MGT (ours)            & 81.6 & 76.1 & 55.5 & 26.4 & 18.9 & 8.0 & 4.6 & 21.2 \\
    \bottomrule
\end{tabular}
    }
\caption{
    \textbf{Quantitative results on \emph{DressCode-test}.}
    Comparing TryOffDiff(single-category) to MGT(unified model).
}
\label{tab:vtoff_dresscode}
\end{table}

No prior VTOFF model handles multiple garment categories. Therefore, to establish a baseline and to measure the cost of unified modeling, we train three category-specific TryOffDiff variants on DressCode (upper-body, lower-body, dresses) and compare them against MGT, a single model conditioned on garment type. This setup tests whether supporting multiple categories in a unified architecture results in performance degradation.
\Cref{tab:vtoff_dresscode} shows MGT achieves comparable performance across all categories, with minor trade-offs (\eg, DISTS: 22.2 vs. 21.6 for upper-body, LPIPS: 30.8 vs. 30.0 for lower-body, and FID: 18.9 vs. 18.4 for dresses). This shows that a single, unified approach can replace multiple specialized models without significant performance loss.

\begin{table}[t]
\setlength{\tabcolsep}{4pt}
\centering
\resizebox{\linewidth}{!}{
\begin{tabular}{lcccccccc}
    \toprule
    \textbf{Method} & SSIM$\uparrow$ & MS-$\uparrow$ & CW-$\uparrow$ & LPIPS$\downarrow$ & FID$\downarrow$ & FD$^\text{CLIP}$$\downarrow$ & KID$\downarrow$ & DISTS$\downarrow$ \\
    \midrule
    TryOffDiff~\cite{velioglu2024tryoffdiff}         & 79.5 & 70.4 & 46.2 & 32.4 & 25.1 & 9.4 & 8.9 & 23.0 \\
    TryOffAnyone~\cite{xarchakos2024tryoffanyone}    & 71.9 & -    & -    & 17.2 & 25.3 & -   & 2.0 & 21.0 \\
    \textsuperscript{$\ddagger$}MGT (ours)           & 78.1 & 66.0 & 39.1 & 36.3 & 21.9 & 7.0 & 8.9 & 24.7 \\
    \bottomrule
\end{tabular}
}
\caption{\textbf{Quantitative results on \emph{VITON-HD-test}.}
\textsuperscript{$\ddagger$}MGT evaluated in cross-dataset setting.
Baseline values from original papers.}
\label{tab:ablations_vitonhd}
\end{table}

We further assess MGT’s generalization in a cross-dataset setting by evaluating it on VITON-HD, despite being trained exclusively on DressCode. As shown in \Cref{tab:ablations_vitonhd}, MGT outperforms TryOffDiff in FID (21.9 vs. 25.1) and remains competitive across other metrics. This indicates strong generalization despite the domain shift.

\begin{table}
\setlength{\tabcolsep}{2.5pt}
\centering
\resizebox{\linewidth}{!}{
\begin{tabular}{l|ccc|ccc}
    \toprule
    & \multicolumn{3}{c|}{VITON-HD} & \multicolumn{3}{c}{DressCode} \\
    \cmidrule(lr){2-4} \cmidrule(lr){5-7}
    Methods & FID$\downarrow$ & FD$^\text{CLIP}$$\downarrow$ & KID$\downarrow$ & FID$\downarrow$ & FD$^\text{CLIP}$$\downarrow$ & KID$\downarrow$ \\
    \midrule
    CatVTON \cite{chong2024catvton}           & 12.0 & 3.5 & 3.9 & 8.4 & \textbf{1.9} & 3.1 \\
    OOTD \cite{xu2024ootdiffusion} + Ground truth         & \textbf{10.8} & \textbf{2.8} & \textbf{2.0} & \textbf{7.5} & 3.4 & \textbf{2.5} \\
    OOTD \cite{xu2024ootdiffusion} + MGT (ours)& \underline{11.9} & \underline{3.3} & \underline{2.6} & \underline{7.9} & 3.4 & \underline{2.7} \\
    \bottomrule
\end{tabular}
}
\caption{\textbf{Quantitative results for p2p-VTON.} 
We evaluate OOTDiffusion (OOTD) with ground truth (GT) garments and MGT-predicted garments, alongside CatVTON, a specialized p2p-VTON model. Our model's output, when integrated with a VTON model, achieves competitive performance, even though it is not explicitly trained for the p2p task.}
\label{tab:p2p}
\end{table}

In p2p-VTON setup~(\cref{tab:p2p}), we pair MGT with OOTDiffusion and compare the results to CatVTON, a dedicated p2p-VTON model. On VITON-HD, OOTD+MGT slightly outperforms CatVTON. On DressCode, MGT achieves competitive results, especially in perceptual metrics, showing its effectiveness in tasks beyond its training objective.

These findings highlight MGT’s versatility across both VTOFF and p2p-VTON tasks. Its improvements in perceptual quality and structural fidelity translate to practical benefits in realistic garment rendering.

\begin{figure}[b]
  \centering
  \begin{subfigure}{0.49\linewidth}
    \includegraphics[width=\linewidth]{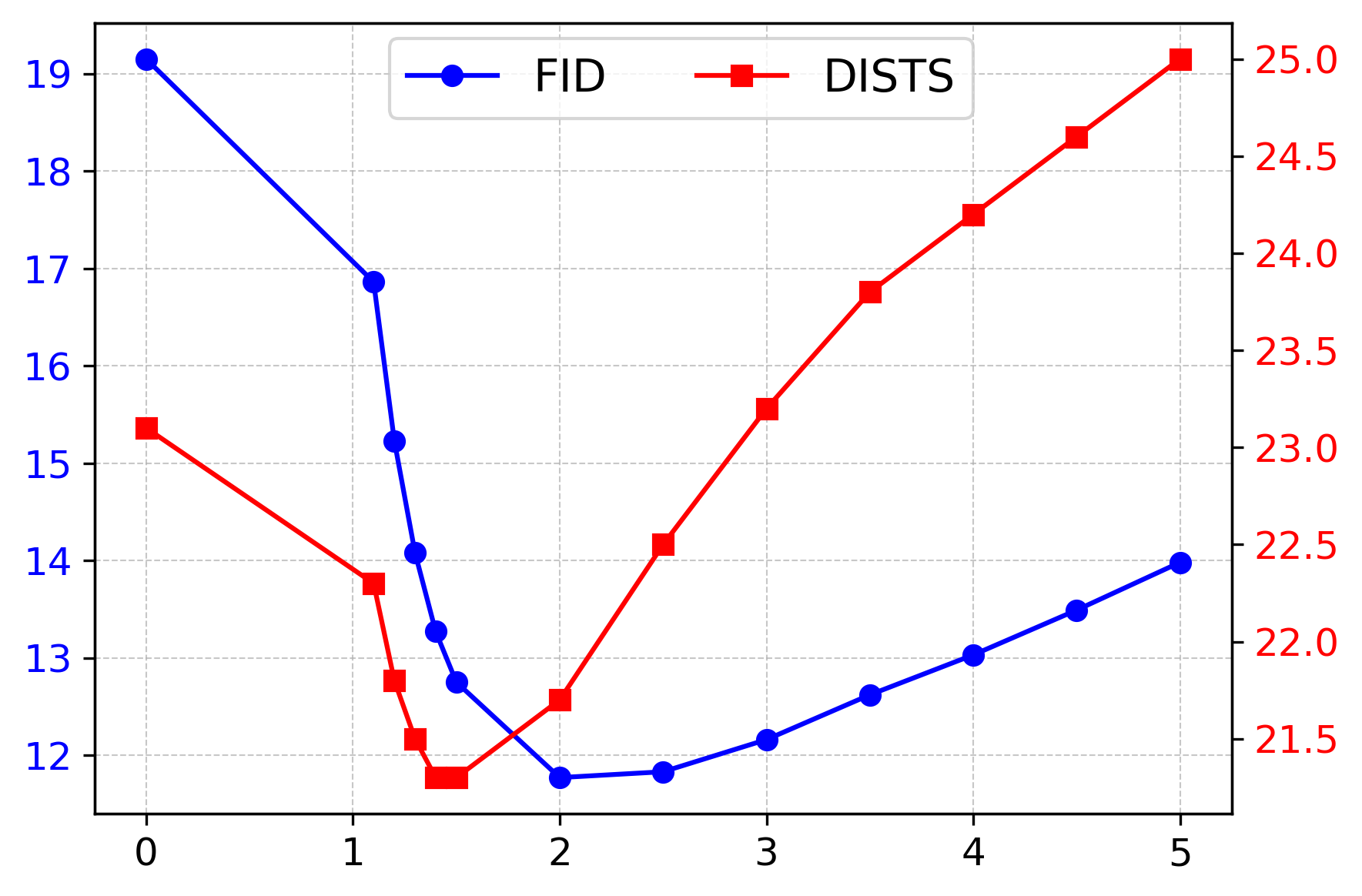}
    \caption{Guidance Scale}
    \label{fig:impact_of_g_n-a}
  \end{subfigure}
  \hfill
  \begin{subfigure}{0.49\linewidth}
    \includegraphics[width=\linewidth]{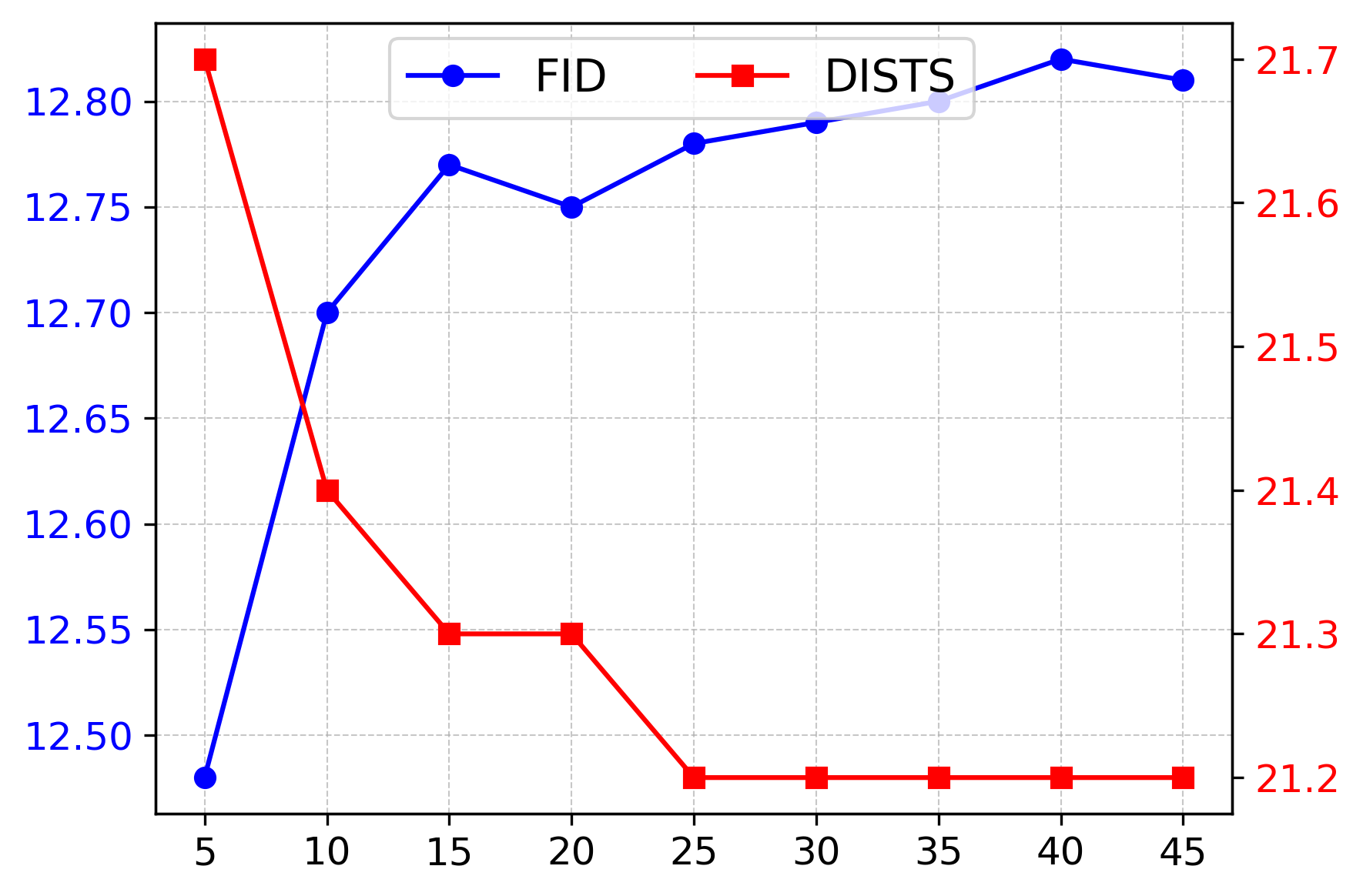}
    \caption{Inference steps}
    \label{fig:impact_of_g_n-b}
  \end{subfigure}
  \hfill
  \caption{\textbf{Impact of guidance scale~($s$) and inference steps~($n$) on DISTS and FID scores.} Evaluated on \emph{DressCode-test} with MGT using the Euler scheduler~\cite{karras2022elucidating}.}
  \label{fig:impact_of_g_n}
\end{figure}

Lastly, \Cref{fig:impact_of_g_n} explores the influence of inference hyperparameters (guidance scale and number of steps) on FID and DISTS scores. This analysis provides insights into optimal inference configurations for MGT.

\begin{figure*}[t]
  \centering
  \includegraphics[width=0.99\linewidth]{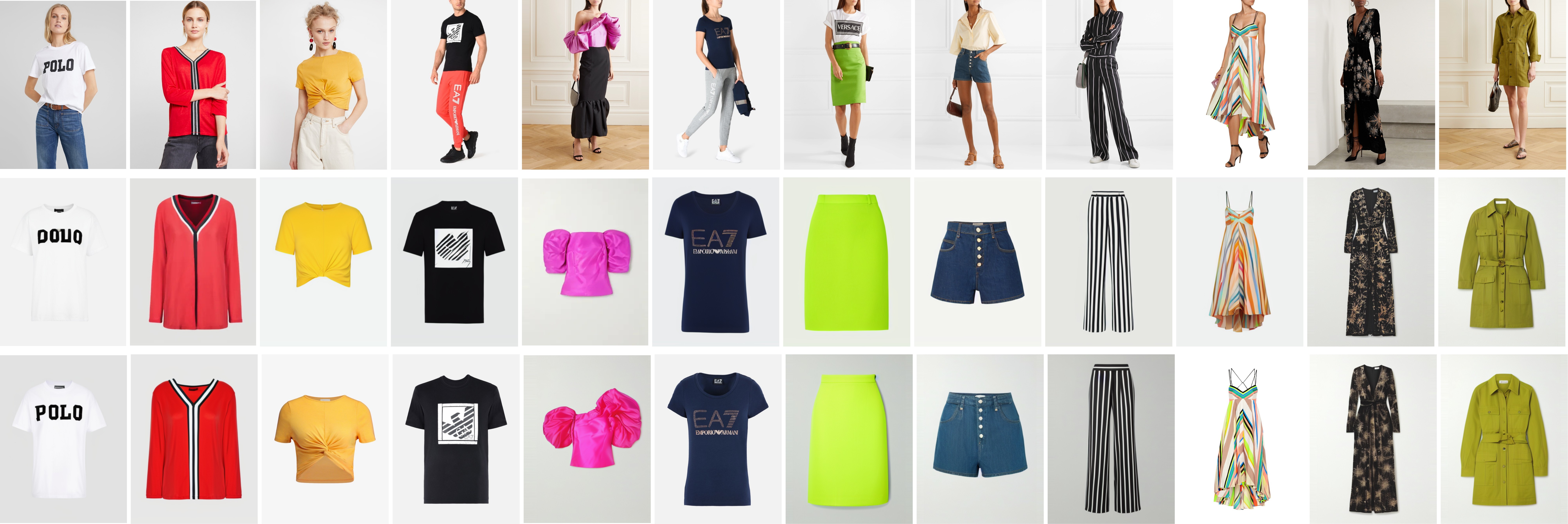}
  \caption{\textbf{Qualitative comparison of garment-specific and Multi-Garment TryOffDiff.} 
  First row displays the reference image.
  Second row shows dataset-specific reconstructions produced by
  TryOffDiff model trained on single category: 
  VITON-HD (cols 1-3),
  DressCode upper-body (cols 4-6), 
  lower-body (cols 7-9),
  and dresses (cols 10-12).
  Last row shows predictions of Multi-Garment TryOffDiff trained on
  full DressCode with garment type conditioning.
  }
  \label{fig:comparison1}
\end{figure*}

\begin{figure}[t]
  \centering
  \begin{subfigure}{0.19\linewidth}
    \includegraphics[width=\linewidth]{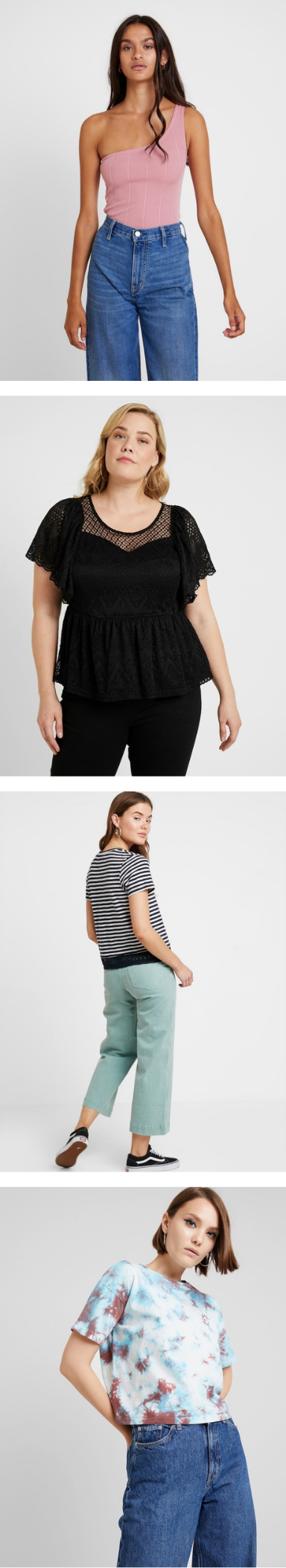}
    \caption{Input}
  \end{subfigure}
  \hfill
  \begin{subfigure}{0.19\linewidth}
    \includegraphics[width=\linewidth]{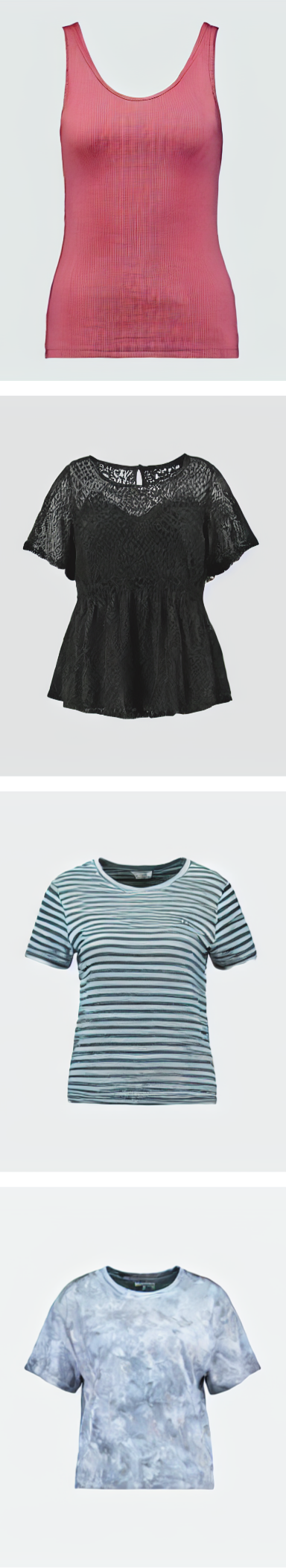}
    \caption{ToD~\cite{velioglu2024tryoffdiff}}
  \end{subfigure}
  \hfill
  \begin{subfigure}{0.19\linewidth}
    \includegraphics[width=\linewidth]{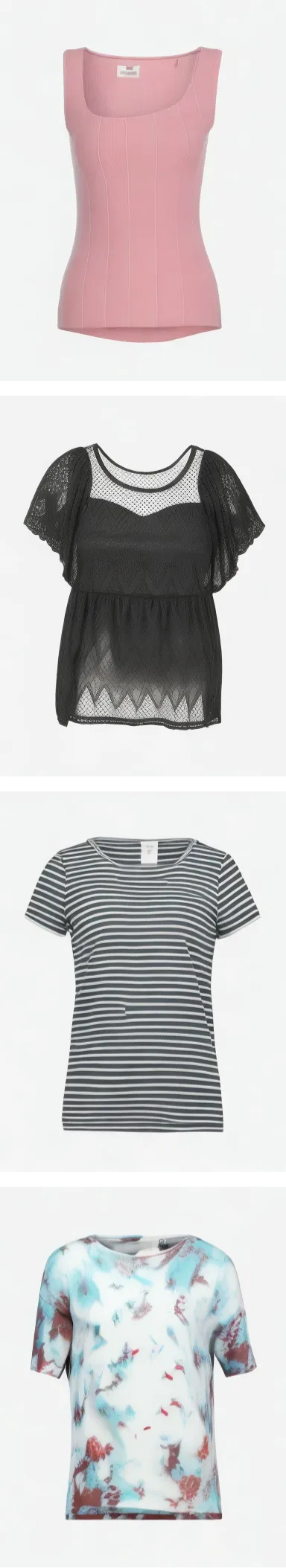}
    \caption{ToA~\cite{xarchakos2024tryoffanyone}}
  \end{subfigure}
  \hfill
  \begin{subfigure}{0.19\linewidth}
    \includegraphics[width=\linewidth]{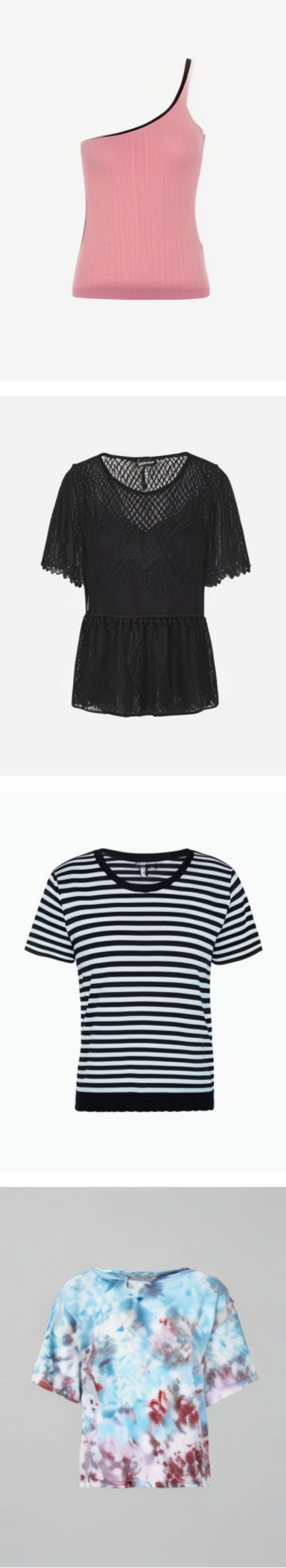}
    \caption{MGT}
  \end{subfigure}
  \hfill
  \begin{subfigure}{0.19\linewidth}
    \includegraphics[width=\linewidth]{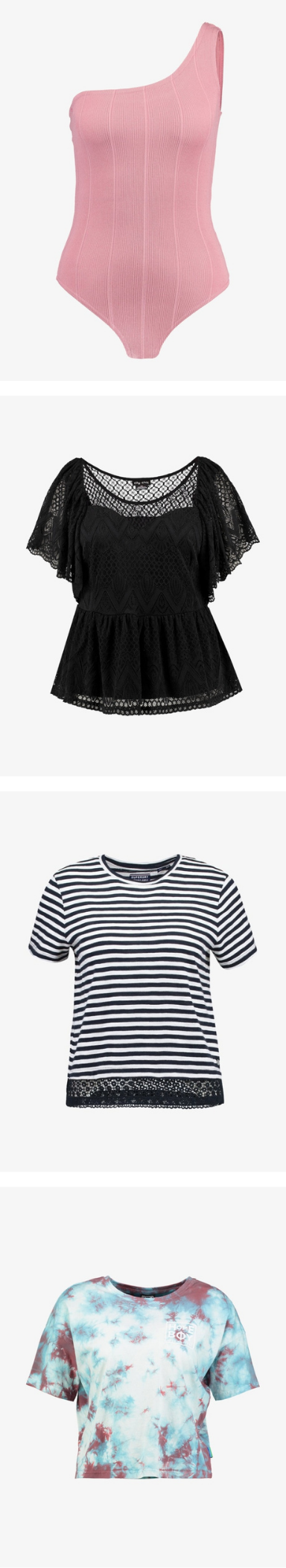}
    \caption{GT}
  \end{subfigure}
  \hfill
  \caption{\textbf{Qualitative comparison on VITON-HD.}
  Despite not being trained on VITON-HD, MGT achieves competitive results compared to baselines specifically trained on it: TryOffDiff~(ToD)~\cite{velioglu2024tryoffdiff} and TryOffAnyone~(ToA)~\cite{xarchakos2024tryoffanyone}.}
  \label{fig:comparison2}
\end{figure}

\subsection{Qualitative Analysis}\label{sec:qualitative}

\Cref{fig:comparison1} shows outputs from two sets of models: category-specific TryOffDiff variants (row 2) and our unified MGT model (row 3). Top row shows input images drawn from; VITON-HD (cols 1–3), DressCode upper-body (4–6), lower-body (7–9), and dresses (10–12).
While all models produce visually plausible outputs, MGT matches or surpasses single-category models in several cases. It maintains clear text (cols 1, 6), emblems (col 4), and accurate garment shapes (col 5), and recovers detailed textures (cols 2, 10). MGT also performs strongly on the unseen VITON-HD dataset (cols 1–3), demonstrating robust generalization across garment types and domains.

To complement the quantitative results on the VITON-HD (\cref{tab:ablations_vitonhd}), we provide a qualitative comparison in \Cref{fig:comparison2}. Despite not being trained on VITON-HD, MGT produces outputs that are visually competitive with baselines that were trained specifically on this dataset, including TryOffDiff~\cite{velioglu2024tryoffdiff} and TryOffAnyone~\cite{xarchakos2024tryoffanyone}. It maintains structural coherence and texture detail, even under challenging conditions involving complex patterns or non-frontal poses. These findings affirm MGT’s ability to generalize without the need for per-category or per-dataset tuning.

\begin{figure*}
  \centering
  \begin{subfigure}{0.49\linewidth}
    \includegraphics[width=\linewidth]{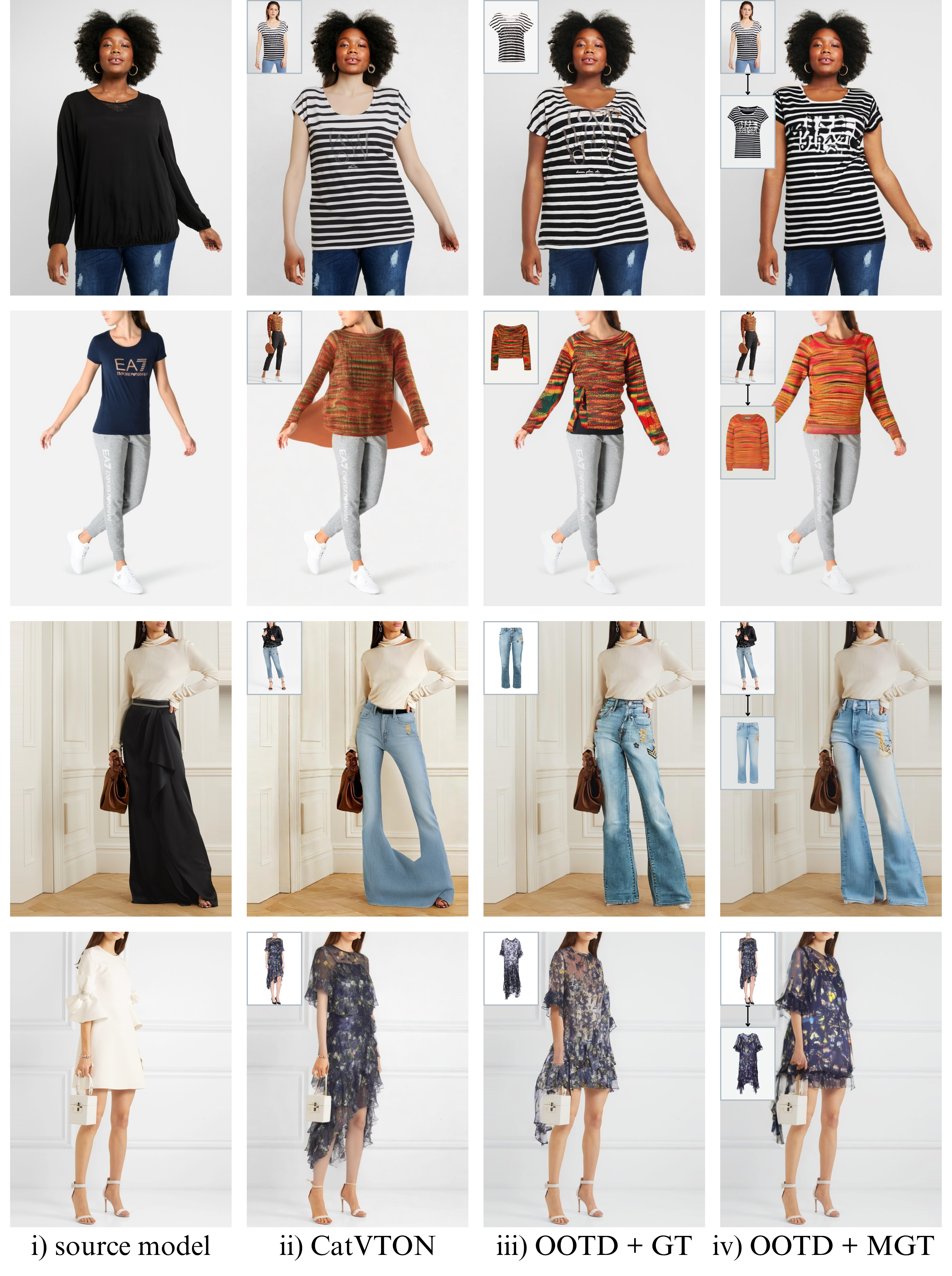}
    \caption{Single Garment}
  \end{subfigure}
  \hfill
  \begin{subfigure}{0.49\linewidth}
    \includegraphics[width=\linewidth]{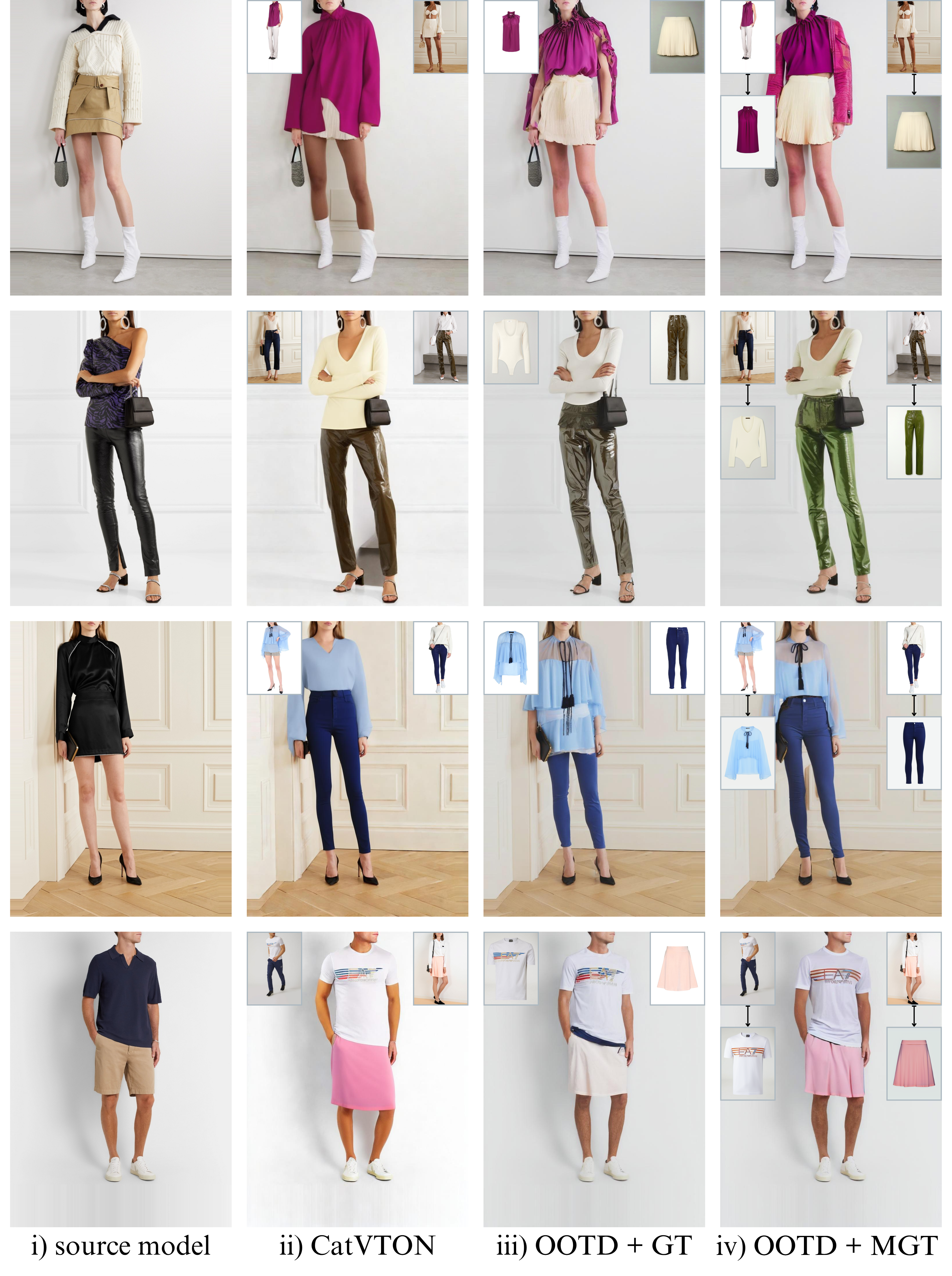}
    \caption{Multi-Garment}
  \end{subfigure}
  \hfill
  \caption{\textbf{Qualitative comparison for Person-to-Person Virtual Try-On in single and multi-garment settings.} 
  Columns: (i) source model image (person to be dressed), (ii) output from CatVTON, using a person image with target garments as condition for direct p2p-VTON, (iii) output from OOTDiffusion~(OOTD), conditioned on the ground-truth~(GT) target garments, and (iv) output from OOTDiffusion with our Multi-Garment TryOffDiff~(MGT) pipeline.
  Each result shows the generated outfit on the source model, with the corresponding target garments shown in the top-left and top-right corners. For single-garment settings (a), only the top-left garment is applied. For multi-garment settings (b), garments are applied sequentially: the top-left garment is applied first, and its output serves as input for applying the top-right garment.}
  \label{fig:comparison3}
\end{figure*}

\Cref{fig:comparison3} shows results for p2p-VTON. CatVTON transfers textures and patterns accurately but occasionally introduces artifacts (rows 1–2) or unintended changes such as skin tone shifts (row 1). In contrast, combining OOTDiffusion with MGT yields consistent, artifact-free outputs. Although no method dominates in all scenarios, the decoupling of garment reconstruction (via MGT) from rendering (via VTON) results in more interpretable and stable outputs.


\section{Conclusion}
\label{sec:conclusion}

We introduced \textbf{Multi-Garment TryOffDiff (MGT)}, a unified latent diffusion model for the Virtual Try-Off (VTOFF) task. By incorporating a lightweight class-conditioning mechanism, MGT supports the reconstruction of diverse garment types—upper-body, lower-body, and dresses—from a single reference image.
Unlike prior work, which relies on separate models for each garment category, MGT achieves near-specialized performance within a single framework. On the DressCode dataset, it matches or surpasses category-specific TryOffDiff variants while supporting multiple categories simultaneously. To our knowledge, MGT is the first unified model to address multi-garment VTOFF.
Moreover, MGT demonstrates strong cross-dataset generalization. Despite being trained exclusively on DressCode, it achieves competitive results on VITON-HD, highlighting its robustness to domain shifts and unseen garments.
Beyond VTOFF, we show that MGT can be seamlessly integrated into person-to-person Virtual Try-On (p2p-VTON) pipelines. When paired with OOTDiffusion, it achieves performance on par with specialized p2p models like CatVTON, while offering a modular and interpretable pipeline, separating garment reconstruction from person rendering.

\textbf{Limitations and Future Work.} MGT currently supports only three garment categories and does not handle layered clothing, constrained by the available dataset annotations. Fine-grained texture recovery and logo preservation also remain challenging, particularly for complex garments. Future research could focus on expanding garment coverage, refining conditioning mechanisms, and exploring higher-capacity architectures or perceptual training objectives to improve detail fidelity.

In summary, MGT marks a step toward general-purpose virtual try-off by unifying garment reconstruction across categories in a single, adaptable model. Its strong performance, generalization capabilities, and modularity make it a valuable component for future virtual fashion systems.

\section*{Acknowledgment}
{This work has been funded by Horizon Europe program under grant agreement 101134447-ENFORCE, and by the German federal state of North Rhine-Westphalia as part of the research funding program KI-Starter. We would like to thank UniZG-FER for providing access to their hardware.}

{
    \small
    \bibliographystyle{ieeenat_fullname}
    \bibliography{main}
}

\end{document}